Technical Memo for Publishing Data in CDVL

# IUPUI Driving Video/Image Benchmark under All Weather and Illumination Conditions

Jiang Yu Zheng, Professor

jzheng@iupui.edu

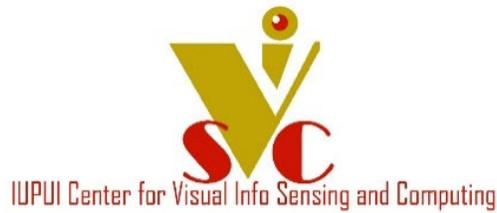

**Center of Visual Information Sensing and Computing**

**School of Science**

**Indiana University Purdue University Indianapolis**

# 1. Introduction

Implementing tasks of public safety often involves driving a vehicle, whether that vehicle is a patrol car for law enforcement or an emergency vehicle at an accident site. Vehicle borne cameras record scenes and events along the routes for real-time driving assistance, road status monitoring, evidence searching, accident analysis, event investigation, etc. Computer vision for driving tasks must understand road conditions regardless of the weather, time, season, and location. Current vision algorithms and tools have not been able to manage all the scenarios of automated driving. Many of the developed sensing algorithms were tested in good weather and under ideal illumination conditions. The development of autonomous driving has moved away from the computer vision approach to other sensors such as LiDAR because of computer vision's failure in adverse weather and under poor illumination conditions. Many of the developed computer vision methods have not covered a wider spectrum of weather due to a lack of data sampled from actual driving. Some of the tests conducted at night even had accidents that caused fatal injury and death. Autonomous driving tests carried out by the automotive industry can limit the vehicle testing on special roads, routes, and areas. However, for the sake of public safety, poor weather and abnormal illumination conditions cannot be ignored by first respond vehicles, whether the task is autonomous driving or evidence recording. A systematic collection and comprehensive analysis of visual data related to a variety of weather categories have to be carried out to make computer vision more robust in driving assistance, especially because cameras are still less expensive than LiDAR devices and computer vision is similar to our human perception of driving a vehicle. For scene and event recording by in-car cameras, various defects or side effects caused by abnormal illumination or poor visibility need to be removed or deducted in target detection and event analysis.

To further extend the robustness of vision-based methods in detecting roads, avoiding collisions, and tracking targets, this project collects sample views from in-car cameras under different illumination and road conditions when public safety vehicles are on patrol and responding to disasters. For intelligent vehicles performing autonomous or assistant driving, the view collection and clustering as images and videos is based on the data mining of naturalistic driving videos [1]. These views, as shown in some examples in Fig. 1, include special appearances that have not yet been completely covered by computer vision in road understanding, vehicle interaction, and collision avoidance, as well as evidence recording. Under different illuminations and media in the atmosphere, our driving views contain snow, rain, direct light, dim lit conditions, sunny facing the sun, shadow, night, and their caused phenomena such as wet roads, glass reflection, glass icing, raining and dirty windshields, moving wipers, etc. in the recording of evidence, accidents, and events. Such IUPUI images and videos created in this project as benchmarks are refined with labels for evaluating algorithms in road detection and navigation, as well as for the training set in machine learning of road environments. The datasets contain 300 5-second HD video clips of driving and 1169 high resolution images on roads covering almost all spectrums of weather and illumination conditions.

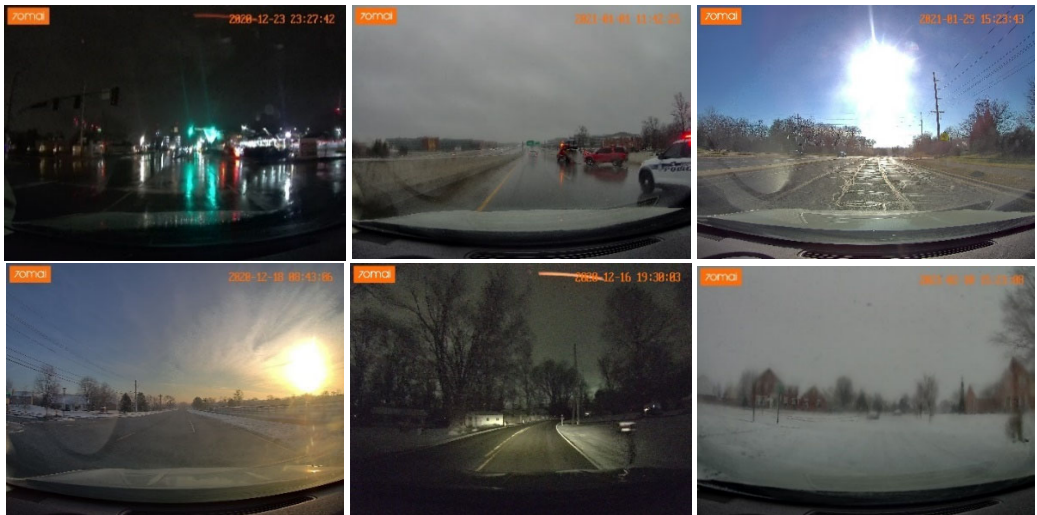

Fig. 1 Some examples in adverse weather and illumination conditions.

We are submitting these video clips in MPEG to the CDVL database, and high-resolution images from an advanced camera as benchmarks with all attributes categories. The datasets contain 1.53GB for videos and 5.65GB for images, respectively. All the video clips and photos were captured during Nov. 2020 and Feb. 2021 by VISC center and they are zipped for publishing in CDVL. All the videos and images are annotated with camera, weather, illumination, and road in their file names. The details of the datasets are further explained in the following sections.

The collected datasets will enhance the computer vision functions in understanding, learning, and testing various scenarios on roads. With the collected dataset covering the spectrum of driving appearances, various algorithms can be developed for different weather and illumination conditions. Special algorithms will be triggered in adverse weather such as snow and night, as well as poor road conditions such as snow covering and wet road surfaces for driving assistance to first responders and routine drivers. In particular, as a compact testing set for computer vision and recognition algorithms, the datasets cover a large variation of road environments at different times. The driving videos captured 5 second increments in order to contain sufficient motion for decision making in the vehicle control and for dynamic event recording. The developed systems can assist law enforcement, emergency vehicles, and the general public for safe driving under different weather and illumination conditions.

# 2. Cameras for Driving Footage

Recent driving cameras have advanced largely in sensibility, resolution, and usability. Table 1 lists cameras used in data capturing. We purchased cameras that have 2.5K resolution and Full-HD (1080p). The cameras use advanced SONY image sensing chips to enhance image intensity for brighter images and video in poor weather and illumination conditions. The cameras also have high dynamic range function (HDR) for high contrast scenes such as evening streetlights. However, the effective field of view for driving is horizontally wide in order to capture different events around, and vertically narrow— excluding vehicle front bonnet and large sky on top in the video. When we acquire a large resolution video, top and bottom parts in the views are excluded to save the data size. Thus, the images and videos in this driving dataset are in the size of 2592×1944 pixels (5MP, 2.5K) and 1280×720 pixels (HD, 1K), respectively, although we still keep the source video clips of 2592×1944 pixels (5MP, 2.5K).

After examining the images and videos, we found that the SONY chip in the camera has an error on intensity overflow if HDR mode is on for video capturing at night. Nevertheless, this hardware defect is insignificant because it appears only at strong lights in the dark evening scenes when the overall exposure of camera is automatically adjusted to the dark background in the large field of view.

The camera is mounted close to the back mirror in a car. The camera roll is one degree counterclockwise with respect to the horizon, and default image is thus equivalent to 1-degree rotated counter-clockwise. The horizon is at the height of 1118 pixels in the video frame from the top, and the EOF (focus of expansion) of vehicle forward translation is measured at (1313, 1118) pixels by using vanishing points in multiple images, where the origin of coordinates system is at top-left corner of the frame. This fixed camera setting is feasible in depth and velocity sensing for the vehicle path planning, feedback control, collision avoidance, and potential danger alert.

Table 1 Cameras used in this view capturing

|  | Camera Spec | Capturing condition |
|---|---|---|
| Camera | 70mai 2K Car Camera 1944p, **Smart Dash Cam Pro** 2.5K, Sony IMX335 2592x1944, WiFi Dash Camera for Cars, Parking Monitor, 2" LCD Screen, Night Vision, iOS/Android Mobile App WiFi, Voice Control (2021), 170 degree wide angle field of view. | 2592x1944 pixels (2.5K), WDR (wide dynamic range) is on. NTSC 30fps. The HDR images are taken by the camera at high contrast scenes when night vision is on. |

# 3. Categorizing Driving Images and Videos

The categorization of driving views under different weather and illumination conditions have been studied in [1]. The files are named by key word attributes to categorize the properties of images and videos captured on the road in driving. The file names are in the form of attributes as

*category1.category2.category3.category4.number*

where each category has two to ten different values or items to express the variation and their combination.

Category 1 is the type of camera used for image or video. P represents an image or photo, and V represents video. The attached number after P or V indicates the camera. In the CDVL, this category may be have been reset to IUPUI.

Category 2 mainly divides different weather conditions, which is the atmosphere of air or additional media of rain, snow, and fog. In the air-only case, we separate views to sunny, cloudy, and night cases, which have point illumination, ambient illumination, and no global illumination. The **sunny** and **cloudy** labels are determined by the local lit area in front of vehicles including on-road and off-road regions, instead of the sky in the distance. For example, if the close area is directly under clouds but the sky in the distance is sunny, the view is still classified as partly cloudy or overcast. On the other hand, if local area is directly under sunlight but the sky in the distance appears cloudy, the view is classified as sunny.

**Night** is classified when the sky is sufficiently dark and both vehicle headlights and street lights are on, and the road ahead of the vehicle has light spots visible from street lights or ego-vehicle headlight. Occasional headlights of opposite vehicles are not definitively classified as night. Depending on the camera, the new camera can increase the exposure or even create high dynamic range images (HDR) for better sensitivity.

**Raining and Snowing** is classified when either rain drops cover the vehicle windshield or the wiper is on. The same classification is applied to snowing as well. The wipers' movement has to be removed in the image according to their shapes or skipped in the video according to their frequency.

**Foggy** is rare in many areas in this season except for some coastal regions. If the vehicle does not move at a fast speed, the camera autoexposure function has been able to increase the contrast to obtain clear views up to a certain distance, e.g., 30m, even during a very foggy morning.

Category 3 is mainly for illumination directions and intensity marked in Bold in Table 2. Depending on the sun's direction as shown in Fig. 2, the illumination direction with respect to the vehicle's direction has values of back to the sun, side to the sun, facing the sun, direct to the sun, and the sun is under the horizon. These sun positions yield nice back to the sun, shadow, facing the sun, direct light, dark lit road areas respectively, whose visual phenomena are further described in Table 3.

Sunny **facing the sun** is classified only when the sun's position is in the camera view, which happens frequently on sunny days when a vehicle is moving south. The visible scenes are either lit or self-shadowed. The road surface may have highlight spots partially due to a wet or pitch repaired surface. While the **direct light** appears during sunrise and sunset when the sun position is close to the center of view, the bright sun may suppress the surrounding scenes and roads due to the limited camera dynamic range. A wide-angle lens camera, such the one we recently purchased with 170 degree may capture the sun more frequently. However, we still consider the direct light as the sun near the center of image.

**Shadow** refers to the condition in which the road ahead has clear shadowy regions and lit regions. In some circumstances, the ground is covered in shade by nearby buildings, bridges, trees, etc.

The **Dark** category refers to the cases of sunny darkness where the nearby ground is not under direct sunlight because it is dusk or before sunrise, when the sun under the horizon. Thus, scenes are dark either in self-shadow or in a shadow cast from other scenes, while the sky is bright due to the reflected sky light from the sun behind the scenes. If the sun is at a central area of the view, this becomes the case of **direct light**. The **dark** illumination also include the dark clouds before storm or close to night under insufficient ambient light.

Table 2 Categories of image and video

| Category 1 | | Catergory 2 | Category 3 | | Category 4 |
|---|---|---|---|---|---|
| Camera | Median | Weather | Illumination | Visual phenomena | Road, Lane, Grass |
| P: image V: video N: cam number. For example P2, V1, V2 | Air – no median | (su) Sunny | (SD) Shadow | Shadow of trees, building, bridge, etc. Dynamic range may not catch changes fast enough | (h) Highway - multiple lanes, ramp, etc. (r) Residential - no center lane mark (s) Suburban - local road with one or two lanes and many vehicles (f) Farm, forest, – narrow roads without lane (u) Urban – building, parked vehicle, and streetlight on sides (p) Parking lot (w) Wet – with mirror reflection on road (c) Snow covered – road feature such as edge and lane mark invisible (b) Barrier and construction site – concrete barrier, guardrail, cones |
| | | | (BS) Back to sun | Including partly cloudy where local area is sun-lit nicely. | |
| | Diffused median | | (FS) Facing the sun | Glare, Road Reflection, dirty glass spot and water stain, reflection of dashboard. | |
| | | | (DL) Direct Light | Glare, glass dirty, surrounding invisible due to narrow camera dynamic range | |
| | | | (DK) Dark | Sky bright, blue, yellow, red, but ground and scenes are not lit and dark. | |
| | | (cd) Cloudy | (PC) Partly cloudy | Near ground is in the shadow casted by cloud. | |
| | | | (OC) Overcast | Mirror reflection on wet road | |
| | | | Dark | Dark cloud before thunderstone | |
| | | (nt) Night | (SL) Streetlight | Urban, bright road | |
| | Water as median | | (HO) Headlight only | Rural, pitch dark, noisy after intensity enhance, glare light on opposite vehicle | |
| | | (rn) Rain | Streetlight Headlight only | Mirror reflection on wet road | |
| | Snow as median | | Overcast | Wiper act, rain drop or flow on glass | |
| | | | Dark | In thunderstorm, low visibility, water | |
| | | (sn) Snow | Overcast | Rural, wiper, icing glass, blurring view | |
| | Heavy median | | Dark | | |
| | | | Streetlight | | |
| | | | Headlight | | |
| | | (fg) Foggy | Overcast | Blurring view | |
| | | | Dark | | |

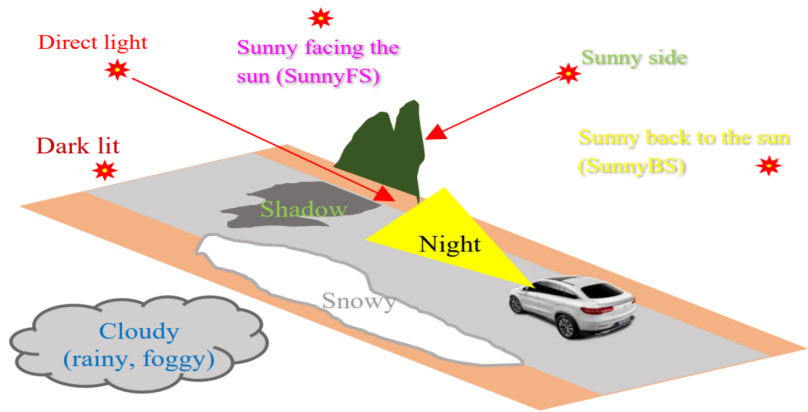

Figure 2 Classified driving views according to light sources, if any, and illumination direction.

VISC, IUPUI, 4/17/2021Cloudy contains three levels as **partly cloudy**, **overcast,** and **dark** (e.g., before thunderstorm) for their intensity. For partly cloudy, we look at surrounding areas on the road to see if objects have a weak shadow, instead of real cloud in the sky miles away because such it does not influence driving environment immediately. The cloud casted shadow has a larger area than tree or building casted shadow.

At night, the road can be illuminated with **streetlights** in city or only by **headlights** of vehicles within dark scenes in suburban areas.

Raining, snowing, and fog cases also have main illuminations as above as bold cases in Table 3 from ambient light of cloud, respectively.

Not every weather category item has a combination with all illumination categories. We placed the illumination categories along with the relevant items of weather category in Table 3. Repeating items that form possible combinations are no longer marked in Bold case.

Category 4 is related to road status and difficulty in the computer vision. The first three difficult cases are **wet** road, **snow-covered** road, and **barriers** on road. A wet road may produce mirror reflection of roadside scenes and lights such that detected features in the road area may not reflect the real shape on the road surface. The snow-covered road features and lane marks are not detectable. We classify the snow-covered roadside in this category as well even if the road area has been cleaned or is melt. The road barrier includes a concrete road divider, construction cones, fences, guardrails, etc. For dry and clear road surfaces excluding wet and snow-covered, we mark them as the following:

**Highway** (h): multiple lanes including ramps.

**Residential** area (r): road without center lane mark including those in some facility, campus, etc.

**Suburban** area (s): general local roads with multiple lanes.

**Urban** (u): buildings, parked cars, street lighting, and pedestrians on roadsides.

**Farm** and forest roads (f): farm or forest road without clear pavement and road edge.

**Parking** lot (p): without clear road definition, many parked vehicle and pedestrians are around.

Three cases of wet, snow covered, and barrier have higher priority to be labeled than other cases because of their rare occurrences, difficulty in recognition, and possible danger they may cause. Table 3 shows the summary of different categories and their relations.

Table 3 A simple layout of classes

| *Weather (media in space)* | *Illumination (light sources, direction, and strength)* | | | | | | | | *Road surface and structure* | | | |
|---|---|---|---|---|---|---|---|---|---|---|---|---|
| Sunny |    |    | DK |    | DL | FS | SD | BS | c |   | b | r, f, h, s, u, p |
| Cloudy |    |    | DK | OC | PC |    |    |    | c | w | b | r, f, h, s, u, p |
| Fog |    |    | DK | OC |    |    |    |    |   |   | b | r, f, h, s, u, p |
| Rain | SL | HO | DK | OC |    |    |    |    |   | w | b | r, f, h, s, u, p |
| Snow | SL | HO | DK | OC |    |    |    |    | c |   | b | r, f, h, s, u, p |
| Night | SL | HO |    |    |    |    |    |    | c | w | b | r, f, h, s, u, p |

The categorization of driving video in terms of weather and illumination conditions is based on our previous research [1] using an old type of cameras. Current research on identifying road and vehicle has largely been focused on sunny and cloudy cases. Lane mark detection has been robust on well-paved roads on cloudy and sunny days. Road edge detection has also been explored on edge detection, region segmentation, and linear feature extraction [2]. Only a few explorations on driving in poor weather have been carried out. Semantic segmentation based on deep learning has also been proposed in recent years to yield road area in rain or snow and dark [5]. More global decisions are based on the layout association of areas trained through labeled samples. Even though, the neural network may fail in rarely trained scenes and scenarios not because they are extremely difficult but due to lack of study on perception and system development so far.

Using the new camera with enhanced sensitivity and wider field of view, several aspects slightly different from our old weather and illumination categories [1] are observed.

(i) The wider angle of view (170 degree vs 120 degree) captures the sun more frequently in the case of sunny facing the sun (su.FS). Therefore, more sunny facing the sun cases are the direct lighting (su.DL) physically as the sunlight reaches the CCD. However, we still label such cases as sunny facing the sun (su.FS) because direct lighting at sun rising or sun set has a larger amount light entering the camera lens in the horizontal direction. This causes the camera exposure adjusted to the sun such that objects and scenes on the ground are suppressed to a dark level, which is different from sunny facing the sun where objects and roads are mostly illuminated brightly.
(ii) Night views by new night vision of the camera are brighter than what was from an old camera even the resulting images have a lot of noises. Also, category Dark (DK) in cloudy case are less possible because images are enhanced by the new cameras.
(iii) Direct light (DL) and Dark (DK) cases are more similar because the glare in direct light case so far may not be as severe as before. The dark parts also show up by high dynamic range (HDR) function of the new camera.
(iv) Dark and Overcast in cloudy, raining, snowing cases with diffuse lighting directions from cloud are less separable using a new camera because it enhances Dark environment. It is considerable to combine Dark in cloudy, raining, snowing (cd.DK, rn.DK, sn.DK) with Overcast (…OC) category. However, Dark in sunny case (su.DK) before sun rise and after sun set is different from those cases. It has Dark ground and bright Sky, and thus should not be merged with Overcast.

Some study has used KNN, Sparse-coding, and decision tree to classify these scenarios using colors sampled in the image frame [2]. Specially sampled areas include sky, off-road, road without vehicle occlusion. The weather and illumination classes have been clustered in advance using unsupervised learning methods such as K-means, and sparse coding [3] based on these data. However, even after weather class identification, road and vehicle detection depends further on road structure and complexity, roadside environments, surrounding vehicles, etc. The most difficult cases for computer vision to understand roads are in adverse weather as

(1) Raining and wet ground (rn.SL.w) where streetlights add additional reflection on the ground, and real road edges are suppressed at signal level by bright lights. In such cases, driving vehicles on a mirror surface is more reliable because off-road materials such as grass and soil are more rough and less reflecting.
(2) Snow covered roads (sn.OC.c, sn.HO.c as well as cd.OC.c) also made the road invisible at various levels. In such an unconstructed road environment, the road recognition is less helpful and the driving task should switch to the recognition of obstacles on roadside.
(3) Night without sufficient illumination except headlight (nt.HO…) may need to confirm the uncertain results detected from weak signals.

The road and vehicle in these weather classes are not able to be modeled only via changing the intensity or contrast in the image. They need to be tackled qualitatively from other angles. For rest of the scenarios, we list general difficult cases or inadequately investigated cases as su.FS, su.DL, rn.HO, etc., which are possible to be modeled using traditional computer vision methods or to avoid their complete modeling according to the driving tasks carried that time. In such cases, some location-independent algorithms can be developed and embedded in the cameras in the future.

# 4. Datasets as Benchmark

## Datasets and Attributes

New cameras with higher resolutions and sensibility as in Table 1 have been purchased in this project for collecting driving images and videos. The data in this set are acquired during Nov. 2020 and March 2021. The camera has wider angle (170 degree) in the field of view and some new functions such as night vision, high dynamic range, and 2.5K resolution (2592x1944pixels). The camera mount is fixed on a single car and is consistent for all dataset. No post-editing has been applied to the raw dataset.

This dataset is ready for publishing with NIST and CDVL. 300 HD Videos (1280x720 pixels) of five seconds are selected in MPEG with the rate of 30 fps. They are representative clips selected from 25hours recorded 2.5K videos in the size of 2592x1944 pixels. Moreover, 1169 high-resolution images with the resolution of 2592x1944 pixels are captured in JPG format. The image and video files are all named in the form of

*category1.category2.category3.category4.number*,

representing camera, weather, illumination, road, and file number, respectively. These attribute values are provided in the file names for machine learning and testing of different weather and illuminations.

Table 4 Number of images and videos for different categories

| Catergory2: 6 levels | Category3: 9 levels | Category4: 9 levels | Category1: 2 levels: (1280×720 for videos), (2592x1944 for images) | |
|---|---|---|---|---|
| Weather | Illumination | Road, Lane, Grass, etc. | V (c2.c3) | P (c2.c3) |
| **(su) Sunny** | **(SD) Shadow** | **(h)** Highway | 17 | 27 |
| | **(BS) Back to sun** | **(r)** Residential | 26 | 89 |
| | **(FS) Facing the sun** | **(u)** Urban | 16 | 81 |
| | **(DL) Direct Light** | **(s)** Suburban | 2 | 4 |
| | **(DK) Dark** | **(f)** Farm and forest | 26 | 67 |
| | | **(p)** Parking lot | | |
| **(cd) Cloudy** | **(PC) Partly cloudy** | (h,…p) | 1 | 29 |
| | **(OC) Overcast** | Snow covered, | 60 | 328 |
| | Dark | wet, etc. | 3 | 31 |
| **(nt) Night** | **(SL) Streetlight** | (h,…, p) | 18 | 50 |
| | **(HO) Headlight only** | | 22 | 69 |
| **(rn) Rain** | Streetlight | **(w) Wet** | 13 | 50 |
| | Headlight only | (h,…p) | 3 | 18 |
| | Overcast | Wiper, rain drop | 11 | 56 |
| | Dark | | 7 | 18 |
| **(sn) Snow** | Overcast | **(c)** Snow **covered** on road or roadside | 23 | 66 |
| | Dark | | | 32 |
| | Streetlight | | 3 | 2 |
| | Headlight | | 13 | 28 |
| **(fg) Foggy** | Overcast | | 23 | 81 |
| | Dark | **(b) Barrier** | 13 | 43 |
| Total | | | **300** | **1169** |

Sample Distribution: We keep the equal distribution of different categories as much as possible in video dataset (300 clips). Images are more random with some categories more than others according to their probabilities of occurrence in the sampling process associated to the Mideast area of the US. Table 4 briefly summarizes the numbers of images and videos in the whole datasets. Figure 3 illustrates the distribution of datasets in terms of different categories.

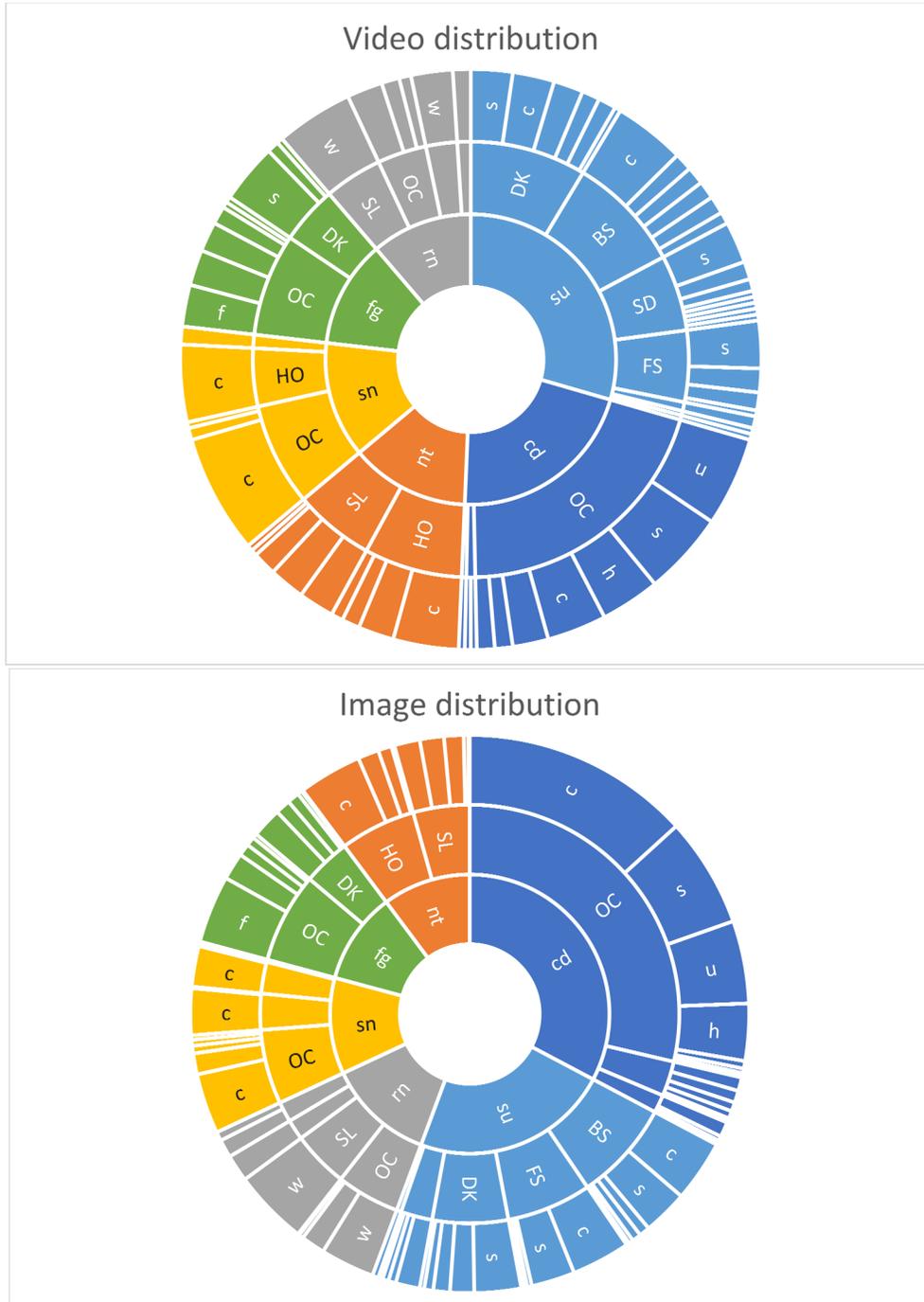

Fig. 3 Sample distributions of datasets for videos and images. Three rings from inside indicate number and percentage of different samples in terms of weather, illumination, and road status.

# 5. File format, Data Access, and Privacy

As proposed in our proposal, 300 mpeg video clips of 5 seconds (30fps, 1280×720) are submitted here to CDVL. The zipped file has 1.53G with a *readme* file in PDF. In addition, 1169 jpg images of 2592x1944 pixels (5.65G in total) are submitted to CDVL. Dataset is captured by the Center of VISC, IUPUI and is provided for publishing with NIST. No post editing has been applied.  No private information such as human faces or home address on mailboxes (no street name and address appear simultaneously) can be identified based on the resolutions of our images and videos. All the sensitive views have been excluded from the datasets.